\documentclass[sigconf, review=False]{acmart}

\usepackage{booktabs}
\usepackage{graphicx}
\usepackage{tabularx}
\usepackage{multirow}
\usepackage{multicol}
\usepackage{enumitem}
\usepackage{caption}
\usepackage{subfig}
\usepackage{amsmath}
\usepackage[ruled,vlined,noend,linesnumbered]{algorithm2e}

\usepackage{titlesec}
\titleformat*{\subsubsection}{\large\bfseries}

\let\oldnl\nl
\newcommand{\nolinenumber}{\renewcommand{\nl}{\let\nl\oldnl}}
\renewcommand{\arraystretch}{1.2}

\newcommand{\MetaLoss}{\mathcal{M}}
\newcommand{\Loss}{\mathcal{L}}
\newcommand{\Task}{\mathcal{T}}
\newcommand{\Transpose}{\mathsf{T}}
\newcommand{\Dataset}{\mathcal{D}}
\newcommand{\Fitness}{\mathcal{F}}

\copyrightyear{2023}
\acmYear{2023}
\setcopyright{acmlicensed}\acmConference[GECCO '23]{Genetic and Evolutionary Computation Conference}{July 15--19, 2023}{Lisbon, Portugal}
\acmBooktitle{Genetic and Evolutionary Computation Conference (GECCO '23), July 15--19, 2023, Lisbon, Portugal}
\acmPrice{15.00}
\acmDOI{10.1145/3583131.3590361}
\acmISBN{979-8-4007-0119-1/23/07}

\begin{document}

\title[Fast and Efficient Local Search for Genetic Programming Based Loss Function Learning]{Fast and Efficient Local Search for Genetic Programming \\ Based Loss Function Learning}

\author{Christian Raymond, Qi Chen, Bing Xue, Mengjie Zhang}
\affiliation{%
  \institution{School of Engineering and Computer Science Victoria University of Wellington}
  \streetaddress{P.O. Box 600, Wellington 6140}
  \city{Wellington}
  \country{New Zealand}
}
\email{christianfraymond@gmail.com}
\email{Qi.Chen, Bing.Xue, Mengjie.Zhang@ecs.vuw.ac.nz}


\begin{abstract}

In this paper, we develop upon the topic of loss function learning, an emergent meta-learning paradigm that aims to learn loss functions that significantly improve the performance of the models trained under them. Specifically, we propose a new meta-learning framework for task and model-agnostic loss function learning via a hybrid search approach. The framework first uses genetic programming to find a set of symbolic loss functions. Second, the set of learned loss functions is subsequently parameterized and optimized via unrolled differentiation. The versatility and performance of the proposed framework are empirically validated on a diverse set of supervised learning tasks. Results show that the learned loss functions bring improved convergence, sample efficiency, and inference performance on tabulated, computer vision, and natural language processing problems, using a variety of task-specific neural network architectures.
  
\end{abstract}


\maketitle

\section{Introduction}

The field of learning-to-learn or \textit{meta-learning} has been an area of increasing interest to the machine learning community in recent years \cite{vanschoren2018meta,peng2020comprehensive}. In contrast to conventional learning approaches, which learn from scratch using a static learning algorithm, meta-learning aims to provide an alternative paradigm whereby intelligent systems leverage their past experiences on related tasks to improve future learning performances \cite{hospedales2020meta}. This paradigm has provided an opportunity to utilize the shared structure between problems to tackle several traditionally very challenging deep learning problems in domains where both data and computational resources are limited and expensive to procure \cite{altae2017low,ignatov2019ai}. 

Many meta-learning approaches have been proposed for optimizing various components of deep neural networks. For example, early research on the topic explored using meta-learning for generating optimization learning rules \cite{schmidhuber1987evolutionary, schmidhuber1992learning, bengio1994use, andrychowicz2016learning}. More recent research has extended itself to learning everything from activation functions \cite{ramachandran2017searching}, fast adaptation parameter initializations \cite{finn2017model, nichol2018first, rajeswaran2019meta}, and neural network architectures \cite{kim2018auto, stanley2019designing, elsken2020meta, ding2022learning} to whole learning algorithms from scratch \cite{real2020automl, co2021evolving} and many more (see \cite{hospedales2020meta}).

However, one component that has been overlooked until very recently is the loss function \cite{wang2022comprehensive}. In deep learning, neural networks are predominantly trained through the backpropagation of gradients originating from the loss function \cite{rumelhart1986learning}; hence, loss functions play an essential role in training neural networks. Given this importance, the typical approach of selecting a loss function heuristically from a modest set of hand-crafted loss functions should be reconsidered, in favor of a more principled data-informed approach that utilizes task-specific information to optimize the selection process.

This paper aims to develop such an approach --- we propose a new framework for loss functions learning called Evolved Model-Agnostic Loss (EvoMAL), which meta-learns symbolic loss functions via a hybrid neuro-symbolic search approach. This new approach unifies two previously divergent lines of research on this topic, which prior to this method, exclusively used either a gradient-based or an evolution-based approach. Furthermore, the proposed framework is both task and model-agnostic, as it can be applied to any technique amenable to a gradient descent style training procedure and is compatible with different model architectures. 

\vspace{-2mm}
\subsection{Contributions}

\begin{enumerate}[leftmargin=*, itemsep=3pt, topsep=3pt]

  \item A promising task and model-agnostic search space composed of primitive mathematical operations and a corresponding genetic programming-based search algorithm are designed for learning symbolic loss functions.
  
  \item Make the first direct connection between expression tree-based symbolic loss functions and gradient trainable loss networks. A procedure for parameterizing and converting the loss functions into trainable loss networks is consequently proposed.
  
  \item Unrolled differentiation, a gradient-based local-search approach is utilized to further optimize the symbolic loss functions. The proposed method is the first computationally tractable approach to further optimizing symbolic loss functions.
  
\end{enumerate}

\section{Background and Related Work}
\label{sec:background}

The field of meta-learning is concerned with the development of self-adapting learning algorithms which learn to solve new tasks more efficiently by utilizing past experiences of solving similar related tasks \cite{hospedales2020meta}. Meta-learning seeks to improve the training dynamics and performance of the final learned model by learning one or more of the inductive biases rather than assuming they are fixed \cite{vilalta2002perspective}. This is achieved by splitting the learning process into two distinct phases \cite{andrychowicz2016learning}, \textit{meta-training} and \textit{meta-testing}.

In the meta-training phase, meta-learning occurs via casting the problem as a bilevel optimization \cite{chen2022gradient}, where the inner optimizaiton uses a set of inner learning algorithms to solve a set of related tasks, minimizing some inner objective. During meta-learning, an outer algorithm updates the inner learning algorithm's inductive biases such that the models learn to improve on some pre-determined outer objective. Following this, in the meta-testing phase, the learned inductive bias is used in standard training to solve a new, unseen but related task.

\subsection{Loss Function Learning}

This paper focuses on one particular form of meta-learning referred to as \textit{loss function learning}. The goal of loss function learning is to learn a loss function $\MetaLoss$ at meta-training time over a distribution of tasks $p(\Task)$. A \textit{task} is defined as a set of input-output pairs $\Task = \{(x_{1}, y_{1}), \dots, (x_{N}, y_{N})\}$, and multiple tasks compose a \textit{meta-dataset} $\Dataset = \{\Task_{1}, \dots, \Task_{M}\}$. Then, at meta-testing time the learned loss function $\MetaLoss$ is used in place of a traditional loss function to train a base learner, e.g. a classifier or regressor, denoted by $f_{\theta}(x)$ with parameters $\theta$ on a new unseen task from $p(\Task)$. In this paper, we constrain the selection of base learners to models trainable via gradient descent style procedures such that we can optimize weights $\theta$ as follows:
\begin{equation}
\theta_{new} = \theta - \alpha \nabla_{\theta} \MetaLoss(y, f_{\theta}(x))
\label{eq:setup}
\end{equation}

\noindent
Several approaches have recently been proposed to accomplish this task, and an  observable trend is that most of these methods fall into one of the following two key categories.
\subsection{Gradient-Based Approaches}

Gradient-based approaches predominantly aim to learn a loss function $\MetaLoss$ through the use of a meta-level neural network external to $f_{\theta}(x)$ to improve on various aspects of the training. For example, in \cite{grabocka2019learning,huang2019addressing}, differentiable surrogates of non-differentiable performance metrics are learned to reduce the misalignment problem between the performance metric and the loss function. Alternatively, in \cite{houthooft2018evolved,bechtle2021meta,collet2022loss,raymond2023online}, loss functions are learned to improve sample efficiency and asymptotic performance in supervised and reinforcement learning, while in \cite{balaji2018metareg,li2019feature,gao2022loss}, they improved on the robustness of a model to domain-shifts and improved domain-generalization.

While the aforementioned approaches have achieved some success, they all have notable limitations. The most salient limitation is that they a priori assume a parametric form for the loss functions. For example, in \cite{bechtle2021meta}, it is assumed that the loss functions take on the parametric form of a two-hidden layer feed-forward neural network with 50 nodes in each layer using ReLU activations. However, such an assumption imposes a bias on the search, often leading to an overparameterized and sub-optimal performing loss function. Another limitation is that these approaches often learn black-box (sub-symbolic) loss functions, which is not ideal, especially in the meta-learning context where \textit{post hoc} analysis of the learned component is crucial, as the intention is to transfer the learned loss function to new unseen problems at meta-testing time.

\subsection{Evolution-Based Approaches}

A promising alternative paradigm is to use evolution-based methods to learn $\MetaLoss$, favoring their inherent ability to avoid local optima via maintaining a population of solutions, their ease of parallelization of computation across multiple processors, and their ability to optimize for non-differentiable functions directly. Examples of such work include \cite{gao2021searching} and \cite{gonzalez2021optimizing}, which both represent $\MetaLoss$ as parameterized Taylor polynomials optimized with covariance matrix adaptation evolutionary strategies (CMA-ES). These approaches generate interpretable loss functions, however; they also assume the parametric form via the degree of the polynomial.

To resolve the issue of having to assume the parametric form of $\MetaLoss$, another avenue of research first presented in \cite{gonzalez2020improved} investigated the use of genetic programming (GP) to learn the structure of $\MetaLoss$ in a symbolic form before applying CMA-ES to optimize the parameterized loss. The proposed method was effective at learning performant loss functions and clearly demonstrated the importance of local search. However, the method had intractable computational costs as using a population-based method (GP) with another population-based method (CMA-ES) resulted in a significant expansion in the number of evaluations at meta-training time, hence it needing to be run on a supercomputer in addition to using truncated training.

Subsequent work in \cite{li2021autoloss} and \cite{liu2021loss} reduced the computational cost of GP-based loss function learning approaches by proposing time-saving mechanisms such as rejection protocols, gradient-equivalence-checking, convergence property verification, and model optimization simulation. These methods successfully reduced the wall-time of GP-based approaches; however, both papers omit the use of local search strategies, which is known to cause sub-optimal performance when using GP \cite{topchy2001faster,smart2004applying,zhang2005learning}. Furthermore, neither method is task and model-agnostic.

\section{Proposed Method}\label{section:c3-proposed-method}

This section presents a detailed description of our new hybrid neuro-symbolic approach named \textit{Evolved Model-Agnostic Loss (EvoMAL)}, which consolidates and extends past research on the topic of loss function learning. The proposed method learns performant symbolic loss functions by solving a bilevel optimization problem. The outer optimization problem involves learning a set of symbolic loss functions, and the inner optimization problem involves performing local search on parameterized versions of the loss functions found in the outer optimization process. To solve this bilevel optimization problem, the evolution-based technique GP is used to solve the discrete outer problem, while unrolled differentiation \cite{wengert1964simple, domke2012generic, deledalle2014stein, maclaurin2015gradient, franceschi2017forward, franceschi2018bilevel, shaban2019truncated, scieur2022curse}, a gradient-based technique previously used in Meta-Learning via Learned Loss (ML$^3$) \cite{bechtle2021meta}, and sometimes referred to as Generalized Inner Loop Meta-Learning (GIMLI) \cite{grefenstette2019generalized}, is used to solve the continuous inner problem. This hybrid learning procedure enables \textit{interpretable} loss functions to be learned on both a lifetime and evolutionary scale.

\subsection{Learning Symbolic Loss Functions}\label{subsec:outer-optimization}

For the outer optimization problem, we propose using GP, a powerful population-based technique that employs an evolutionary search to directly search the set of primitive mathematical operations \cite{koza1992genetic}. In GP, solutions are composed of terminal and function nodes in a variable-length hierarchical expression tree-based structure. This symbolic structure is a natural and convenient way to represent loss functions, due to its high interpretability and trivial portability to new problems. Transferring a learned loss function to new problems often only requires a line or two of additional code.

\subsubsection{Search Space Design}
~\\
In order to utilize GP, a search space containing promising loss functions must first be designed. When designing the desired search space, four key considerations were made --- first, the search space should superset existing loss functions such as the squared error in regression and the cross entropy loss in classification. Second, the search space should be dense with promising new loss functions while also containing sufficiently simple loss functions such that cross-task generalization can occur successfully at meta-testing time without overfitting. Third, ensuring that the search space satisfies the key property of GP closure, \textit{i.e.} loss functions will not cause \textit{NaN}, \textit{Inf}, undefined, or complex output. Finally, ensuring that the search space is both task and model-agnostic. With these considerations in mind, we present the function set in Table \ref{table:function-set}. Regarding the terminal set, the loss function arguments $f_{\theta}(x)$ and $y$ are used, as well as (ephemeral random) constants $+1$ and $-1$.

\begin{table}[t!]
\centering
\captionsetup{justification=centering}
\caption{Task and model-agnostic function set.}
\setlength{\tabcolsep}{15pt}
\begin{tabular}{l|c|c}
\hline
Operation      & Expression                 & Arity \\      \hline \hline
Addition       & $x_1 + x_2$                & 2     \\
Subtraction    & $x_1 - x_2$                & 2     \\
Multiplication & $x_1  \cdot x_2$           & 2     \\
Division ($AQ$)& $x_1 / \big(\sqrt{\smash[b]{1 + x_2^2}}\big)$ & 2     \\ \hline \hline
Square         & $x^2$                      & 1     \\
Absolute       & $|x|$                      & 1     \\
Square Root    & $\sqrt{|x| + \epsilon}$    & 1     \\
Natural Log    & $\ln{(|x| + \epsilon)}$    & 1     \\ \hline
\end{tabular}
\label{table:function-set}
\end{table}
Unlike previously proposed search spaces for loss function learning, we have made several necessary amendments to ensure proper GP closure, and sufficient task and model-generality. The salient differences are as follows: 

\begin{itemize}[leftmargin=*, itemsep=3pt, topsep=3pt]

  \item In \cite{gonzalez2020improved}, the natural log ($\ln(x)$), square root ($\sqrt x$), and division $(x_1/x_2)$ operators were used in the function set. Using these unprotected operations can result in imaginary or undefined output violating the GP closure property. To satisfy the closure property, we replace both the natural log and square root with protected alternatives, as well as replace the division operator with the analytical quotient ($AQ$) operator, a smooth and differentiable approximation to the division operator \cite{ni2012use}.
  
  \item The proposed search space is both task and model-agnostic in contrast to \cite{li2021autoloss} and \cite{liu2021loss}, which use multiple aggregation-based and element-wise operations in the function set. These operations make sense within the respective paper's domains (object detection) but does not make sense when applied to other tasks such as tabulated and natural language processing problems.
  
\end{itemize}

\subsubsection{Outer Search Algorithm Design}
\begin{figure}
\centering

    \includegraphics[width=0.9\columnwidth]{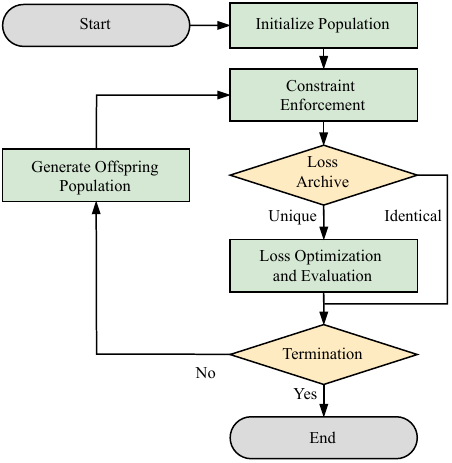}
    \caption{Overview of the EvoMAL algorithm.}
    
\label{fig:flowchart}
\end{figure}
~\\
The search algorithm used in the discrete outer optimization process of EvoMAL uses a prototypical implementation of GP. An overview of the algorithm is as follows:

\begin{itemize}[leftmargin=*, itemsep=3pt, topsep=3pt]

    \item \textbf{Initialization}: To generate the initial population of candidate loss functions, 25 expression trees are randomly generated using \textit{Ramped Half-and-Half}  where the inner nodes are selected from the function set and the leaf nodes from the terminal set. 

\end{itemize}

\noindent
Subsequently, the main loop begins by performing the inner loss function optimization and evaluation process to determine each loss function's respective fitness (discussed in detail in Section \ref{subsec:inner-optimization}). Following this, a new offspring population of equivalent size is constructed via the crossover, mutation, selection, and elitism genetic operators.

\begin{itemize}[leftmargin=*, itemsep=3pt, topsep=3pt]

  \item \textbf{Crossover}: For the crossover genetic operator, two loss functions are selected and then combined via a \textit{One-Point Crossover} with a probability of 70\%, which slices the two selected loss functions together to form a new loss function. 
  
  \item \textbf{Mutation}: For the mutation genetic operator, a loss function is selected from the population, and a \textit{Uniform Mutation} is applied with a mutation rate of 25\%, which in place modifies a sub-expression at random with a newly generated sub-expression. 
  
  \item \textbf{Selection}: Selection of candidate loss functions from the population for crossover and mutation is achieved via a standard \textit{Tournament Selection}, which selects 4 loss functions from the population at random and returns the loss function with the best fitness score.
  
  \item \textbf{Elitism}: To ensure that performance does not degrade throughout the evolutionary process, elitism is used to retain the top-performing loss functions with an elitism rate of 5\%. 
  
\end{itemize}

\noindent
The main loop is iteratively repeated up to 50 times until convergence, and the loss function with the best fitness is selected as the final learned loss function. For clarity, we include an overview of the outer optimization process in Figure \ref{fig:flowchart}.

\subsubsection{Constraint Enforcement}
\begin{figure}[]
\centering

    \includegraphics[width=\columnwidth]{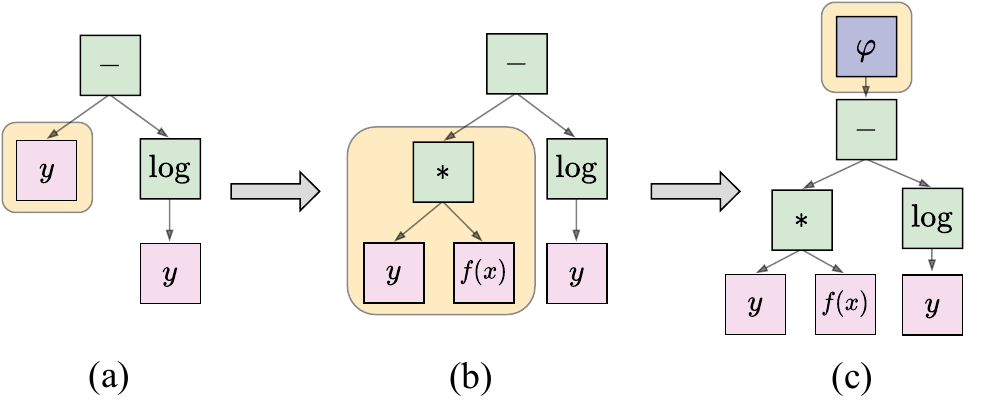}
    \captionsetup{justification=centering}
    \caption{Overview of the constraint enforcement procedure, where (a) is a constraint violating expression, (b) demonstrates enforcing the required arguments constraint, and (c) shows enforcing the non-negative output constraint.}
    
\label{fig:constraints}
\end{figure}

~\\
When using GP, the learned expressions can occasionally violate two important constraints of a loss function. \textbf{(1) Required Arguments Constraint:} A loss function must have as arguments $f_{\theta}(x)$ and $y$ by definition. \textbf{(2) Non-Negative Output Constraint:} A loss function should always return a non-negative output such that $\forall x,\forall y, \forall f_{\theta}[\MetaLoss(y, f_{\theta}(x)) \geq 0]$. To resolve this we describe two corresponding correction strategies, which we summarize in Figure \ref{fig:constraints} for clarity.

\begin{enumerate}[leftmargin=*, itemsep=3pt, topsep=3pt]

  \item \textbf{Required Arguments Constraint:} In \cite{gonzalez2020improved}, violating expressions were assigned the worst-case fitness, such that selection pressure would phase out those loss functions from the population. Unfortunately, this approach degrades search performance, as a subset of the population is persistently searching infeasible regions of the search space. To resolve this, we propose a simple but effective corrections strategy to violating loss functions, which randomly selects a terminal node and replaces it with a random binary node, \textit{i.e.} function node with an arity of 2, with arguments $f_{\theta}(x)$ and $y$ in no predetermined order (required for non-associative operations).
  
  \item \textbf{Non-Negative Output Constraint:} An additional constraint \textit{optionally} enforced in the EvoMAL algorithm is that the learned loss function should always return a non-negative output $\MetaLoss:\mathbb{R}^2 \rightarrow \mathbb{R}_{0}^{+}$. This is achieved via all learned loss function's outputs being passed through an activation function $\varphi$, which was selected to be the smooth $Softplus(x)=ln(1+e^{x})$ activation.
  
\end{enumerate}
\subsubsection{Loss Archival Strategy}
~\\
As computational efficiency is often of concern when using population-based methods, a loss archival strategy based on a key-value pair structure with $\Theta(1)$ lookup is used to ensure that symbolically equivalent loss functions are not reevaluated.

\subsection{Loss Function Optimization and Evaluation}\label{subsec:inner-optimization}

\begin{figure}[t!]
    \centering
    \subfloat[\centering Example learned loss function $\MetaLoss$.]
    {{\includegraphics[width=6cm]{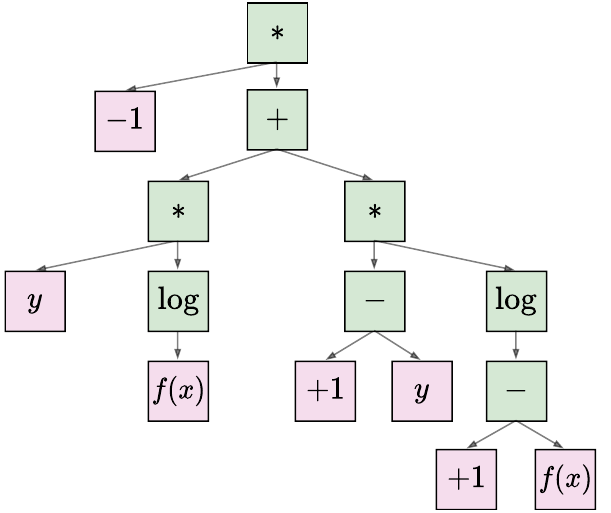}}}
    \qquad
    \subfloat[\centering Example meta-loss network $\MetaLoss_{\phi}^{\Transpose}$.]
    {{\includegraphics[width=7.5cm]{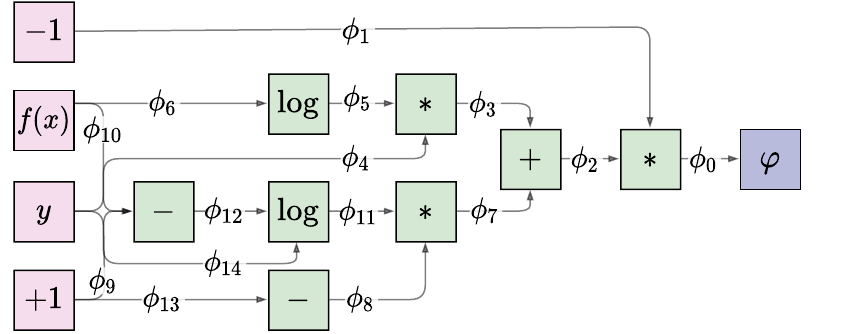}}}
    
    \captionsetup{justification=centering}
    \caption{Overview of the transitional procedure used to covert $\MetaLoss$ into a trainable meta-loss network $\MetaLoss_{\phi}^{\Transpose}$.}%
    \label{fig:phase-transformer}%
\end{figure}

Numerous empirical results have shown that local search is imperative when using GP to get state-of-the-art results \cite{chen2015generalisation, gonzalez2020improved}. Therefore, unrolled differentiation, an efficient gradient-based local search approach is integrated into the proposed method. To integrate ML$^3$ into the EvoMAL framework, we must first transform the expression tree-based representation of $\MetaLoss$ into a compatible network-style representation. To achieve this, we propose the use of a \textit{transitional procedure} that takes each loss function $\MetaLoss$, represented as a GP expression and converts it into a trainable network, \textit{i.e.} a weighted directed acyclic graph, as shown in Figure \ref{fig:phase-transformer}. First, a graph transpose operation $\MetaLoss^{\Transpose}$ is applied to reverse the edges such that they now go from the terminal (leaf) nodes to the root node. Following this, the edges of $\MetaLoss^{\Transpose}$ are parameterized by $\phi$, giving $\MetaLoss_{\phi}^\Transpose$, which we further refer to as a meta-loss network to delineate it clearly from its prior state. 

Finally, to initialize $\MetaLoss_{\phi}^\Transpose$, the weights are sampled from $\phi\sim\mathcal{N}(1, 1\mathrm{e}{-3})$, such that $\MetaLoss_{\phi}^\Transpose$ is initialized from its (near) original unit form, where the small amount of variance is to break network symmetry. For computational efficiency we use an adjacency list representation at implementation level which enables both the transpose and parameterization steps to occur simultaneously with a linear time and space complexity $\Theta(|\mathcal{V}| + |\mathcal{E}|)$ with respect to the number of vertices $\mathcal{V}$ and edges $\mathcal{E}$ (i.e. nodes and weights) in the learned loss function.

\subsubsection{Extension to the Multi-Output Setting}
~\\
To extend this framework to $\mathcal{C}$-way multi-output tasks (such as multi-class classification) we apply the binary version of the loss to each label and then take the sum across the outputs.
\begin{equation}
\MetaLoss_{\phi}^{\Transpose}(y, f_{\theta}(x)) = \sum_{i=1}^{\mathcal{C}}\MetaLoss_{\phi,(i)}^{\Transpose}(y_{(i)}, f_{\theta,(i)}(x))
\label{eq:multi-output}
\end{equation}
\subsubsection{Loss Function Optimization}

\begin{figure*}[]

    \includegraphics[width=\textwidth]{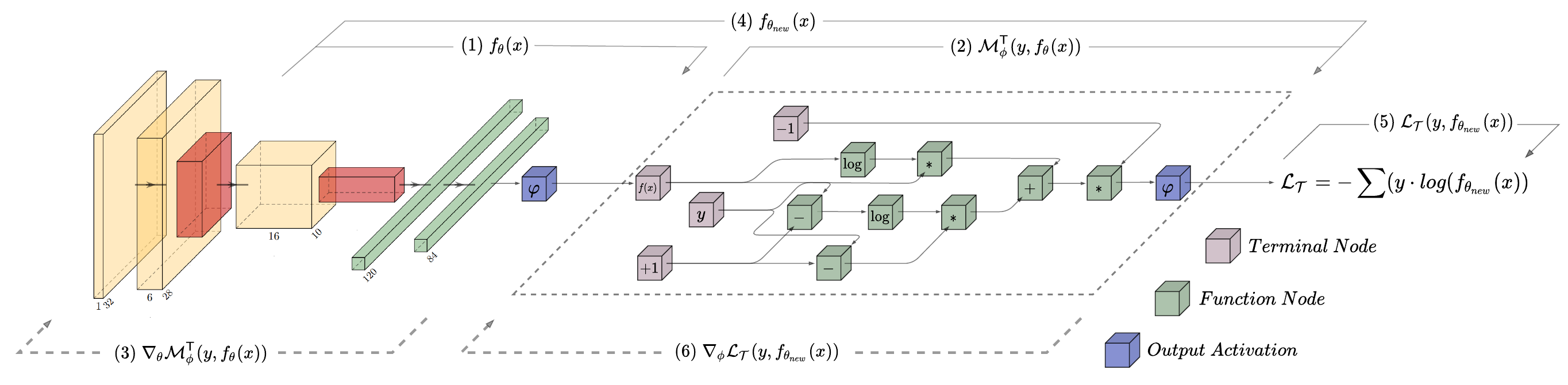}
    \captionsetup{justification=centering}
    \caption{Example of one step of the inner loss optimization process used in EvoMAL to learn the weights $\phi$ of the meta-loss network $\MetaLoss_{\phi}^{\Transpose}$ with respect to the base network $f_{\theta}(x)$ shown (left) as the popular LeNet-5 architecture at meta-training time.}

\label{fig:network}
\end{figure*}

\begin{algorithm}[t!]
\SetAlgoLined
\DontPrintSemicolon
\SetKwInput{Input}{In}
\BlankLine
\Input{
    $\MetaLoss_{\phi}^{\Transpose} \leftarrow$ Loss function learned by GP\newline
    $\mathcal{S}_{meta} \leftarrow$ Number of meta gradient steps\newline
    $\mathcal{S}_{base} \leftarrow$ Number of base gradient steps
}
\nolinenumber\hrulefill
\BlankLine
\For{$i \in \{0, ... , S_{meta}\}$}{
    \For{$j \in \{0, ... , |\Dataset_{Train}|\}$}{
        $\theta \leftarrow$ Initialize parameters of base learner\;
        \For{$k \in \{0, ... , \mathcal{S}_{base}\}$}{
            $X_{j}$, $y_{j}$ $\leftarrow$ Sample task $\Task_{j} \sim p(\Task)$\;
            $\MetaLoss_{learned} \leftarrow \MetaLoss_{\phi}^{\Transpose}(y_{j}, f_{\theta_{j}}(X_{j}))$\;
            $\theta_{new_{j}} \leftarrow \theta_{j} - \alpha \nabla_{\theta_{j}} \MetaLoss_{learned}$\;
        }
        $\Loss_{task_{j}} \leftarrow \Loss_{\Task}(y_{j}, f_{\theta_{new_j}}(X_j))$\;
    }
    $\phi \leftarrow \phi - \eta \nabla_{\phi} \sum_{j} \Loss_{task_{j}}$\;
}
\BlankLine
\caption{Inner Loss Function Optimization}
\label{alg:meta-training}
\end{algorithm}
~\\
For simplicity, we constrain the description of loss function optimization to the vanilla backpropagation case where the meta-training set $\Dataset_{Train}$ contains one task, \textit{i.e.} $|\Dataset_{Train}| = 1$; however, the full process where $|\Dataset_{Train}| > 1$ is summarized in Algorithm \ref{alg:meta-training}. To learn the weights $\phi$ of the meta-loss network $\MetaLoss_{\phi}^\Transpose$ at meta-training time with respect to base learner $f_{\theta}(x)$, we first use the initial values of $\phi$ to produce a loss value $\MetaLoss_{learned}$ based on the forward propagation of $f_{\theta}(x)$.
\begin{equation}
\MetaLoss_{learned} = \MetaLoss_{\phi}^{\Transpose}(y, f_{\theta}(x))
\label{eq:loss-base}
\end{equation}
\noindent
Using $\MetaLoss_{learned}$, the weights $\theta$ are optimized by taking the gradient of the loss value with respect to $\theta$, where $\alpha$ is the base learning rate as shown in Equation (\ref{eq:backward-base}). Note, multiple gradient steps of $\theta$ can be taken here, which requires back-propagating through a chain of $f_{\theta}(x)$ gradient steps; however, in practice we find similar to \cite{bechtle2021meta}, that  
a single gradient step on $\theta$ is sufficient.
\begin{equation}
\begin{split}
\theta_{new}
& = \theta - \alpha \nabla_{\theta} \MetaLoss_{\phi}^{\Transpose}(y, f_{\theta}(x)) \\
& = \theta - \alpha \nabla_{\theta} \mathbb{E}_{\mathsmaller{X}, y} \big[ \MetaLoss_{learned} \big]
\end{split}
\label{eq:backward-base}
\end{equation}
\noindent
This gradient computation can be decomposed via the chain rule into the gradient of $\MetaLoss_{\phi}^{\Transpose}$ with respect to the product of the base models predictions $f_{\theta}(x)$ and the gradient of $f$ with parameters $\theta$.
\begin{equation}
\theta_{new} = \theta - \alpha \nabla_{f} \MetaLoss_{\phi}^{\Transpose}(y, f_{\theta}(x)) \nabla_{\theta}f_{\theta}(x)
\label{eq:backward-base-decompose}
\end{equation}
Following this, $\theta$ has been updated to $\theta_{new}$ based on the current meta-loss network weights; $\phi$ now needs to be updated to $\phi_{new}$ based on how much learning progress has been made. Using the new base learner weights $\theta_{new}$ as a function of $\phi$, we utilize the concept of a \textit{task loss} $\Loss_{\Task}$ (inner objective) to produce a loss value $\Loss_{task}$ to optimize $\phi$ through $\theta_{new}$.
\begin{equation}
\Loss_{task} = \Loss_{\Task}(y, f_{\theta_{new}}(x))
\label{eq:loss-meta}
\end{equation}
where $\Loss_{\Task}$ is selected based on the respective application --- for example, the mean squared error (\small$MSE$\normalsize) loss for the task of regression, binary cross-entropy (\small$BCE$\normalsize) loss for binary classification, and categorical cross-entropy (\small$CCE$\normalsize) loss for multi-class classification.
%
%
Optimization of the meta-loss network loss weights $\phi$ now occurs by taking the gradient of $\Loss_{\Task}$ with respect to $\phi$, where $\eta$ is the meta learning rate.
\begin{equation}
\begin{split}
\phi_{new}
& = \phi - \eta \nabla_{\phi}\Loss_{\Task}(y, f_{\theta_{new}}(x)) \\
& = \phi - \eta \nabla_{\phi} \mathbb{E}_{\mathsmaller{X}, y} \big[ \Loss_{task} \big]
\end{split}
\label{eq:backward-meta}
\end{equation}
where the gradient computation can be decomposed by applying the chain rule as shown in Equation (\ref{eq:backward-meta-decompose}) where the gradient with respect to the meta-loss network weights $\phi$ requires the new model parameters $\theta_{new}$. 
\begin{equation}
\phi_{new} = \phi - \eta \nabla_{f}\Loss_{\Task} \nabla_{\theta_{new}} f_{\theta_{new}} \nabla_{\phi}\theta_{new} 
\label{eq:backward-meta-decompose}
\end{equation}
This process is repeated for a predetermined number of meta gradient steps $S_{meta}$. Following each meta gradient step, the base learner weights $\theta$ is reset. Note that in Equations (\ref{eq:backward-base})--(\ref{eq:backward-base-decompose}) and (\ref{eq:backward-meta})--(\ref{eq:backward-meta-decompose}), the gradient computation can alternatively be performed via automatic differentiation. Figure \ref{fig:network} shows an example of one step of the inner loss optimization process used in EvoMAL to learn the meta-loss network weights $\phi$.

\subsubsection{Loss Function Evaluation}

\begin{algorithm}[t!]
\SetAlgoLined
\DontPrintSemicolon
\SetKwInput{Input}{In}
\BlankLine
\Input{
$\MetaLoss_{\phi}^{\Transpose} \leftarrow$ Loss function learned by EvoMAL\newline
$S_{base} \leftarrow$ Number of base gradient steps
}

\nolinenumber\hrulefill
\BlankLine
\For{$i \in \{0, ... , |\Dataset_{Train}|\}$}{
    $\theta_{i} \leftarrow$ Initialize parameters of base learner $f_{\theta_{i}}$\;
    $X_{i}$, $y_{i}$ $\leftarrow$ Sample task $\Task_{i} \sim p(\Task)$\;
    \For{$j \in \{0, ... , S_{base}\}$}{
        $\MetaLoss_{learned} \leftarrow \MetaLoss_{\phi}^{\Transpose}(y_{i}, f_{\theta_{i}}(X_{i}))$\;
        $\theta_{i} \leftarrow \theta_{i} - \alpha \nabla_{\theta_{i}} \MetaLoss_{learned}$\;
    }
}
$\Fitness \leftarrow \frac{1}{|\Dataset_{Train}|}\sum_{i}\Loss_{\mathcal{P}}(y_{i}, f_{\theta_{i}}(X_{i}))$
\BlankLine
\caption{Loss Function Evaluation}
\label{alg:meta-testing}
\end{algorithm}
~\\
To derive the fitness $\Fitness$ of $\MetaLoss_{\phi}^{\Transpose}$, a conventional training procedure is used as summarized in Algorithm \ref{alg:meta-testing}, where $\MetaLoss_{\phi}^{\Transpose}$ is used in place of a traditional loss function to train $f_{\theta}(x)$ over a predetermined number of base gradient steps $S_{base}$. This training process is identical to training at meta-testing time. The final average inference performance of $\MetaLoss_{\phi}^{\Transpose}$ across all the tasks in $\Dataset_{Train}$ is assigned to $\Fitness$, where any differentiable or non-differentiable performance metric $\Loss_{\mathcal{P}}$ (outer objective) can be used. In our experiments, we assign $\Loss_{\mathcal{P}}$ to be the $\small MSE\normalsize$ for regression, and error rate ($\small ER \normalsize$) for classification.
\begin{equation}
\Fitness = \frac{1}{|\Dataset_{Train}|}\sum_{i}\Loss_{\mathcal{P}}(y_{i}, f_{\theta_{i}}(X_{i}))
\label{eq:fitness}
\end{equation}

\section{Experimental Design}\label{section:c3-experiment-setup}

In this section, we evaluate the performance of EvoMAL for the task of loss function learning. A set of experiments are conducted across four datasets and the performance is contrasted against a representative set of methods implemented in DEAP \cite{fortin2012deap}, PyTorch \cite{paszke2019pytorch} and Higher \cite{grefenstette2019generalized}, which are as follows:
\begin{itemize}[leftmargin=*, itemsep=3pt, topsep=3pt]

    \item \textbf{Baseline} -- Directly using $\Loss_{\Task}$ as the loss function, \textit{i.e.} using loss functions $\small MSE \normalsize$, $\small BCE \normalsize$ or $\small CCE \normalsize$ respectively.
    
    \item \textbf{Random} -- Pure random search on the symbolic search space proposed in Section \ref{subsec:outer-optimization}, where loss functions represented as expression trees are randomly generated with an equivalent number of evaluations to EvoMAL.
    
    \item \textbf{GP-LFL} -- A proxy method used to aggregate previous GP-based approaches for loss function learning \textit{without} any local search mechanisms, using an identical setup to EvoMAL excluding Section \ref{subsec:outer-optimization}.
    
    \item \textbf{ML$^3$ Supervised} -- Gradient-based loss learning method proposed in \cite{bechtle2021meta}, which uses a parametric loss function defined by a two hidden layer feed-forward network trained with the method shown in Section \ref{subsec:inner-optimization}.
    
\end{itemize}
\noindent
The experimental design intends to isolate and highlight the effects of the different components in EvoMAL, to validate the effectiveness of hybridizing existing loss function learning approaches into one unified framework. The code for reproducing all the experiments can be found at: 
\href{https://github.com/Decadz/Evolved-Model-Agnostic-Loss}{https://github.com/Decadz/Evolved-Model-Agnostic-Loss}.

\subsection{Benchmark Problems}

For all benchmark problems, stochastic gradient descent ($\small SGD \normalsize$) is used as the optimizer similar to \cite{gonzalez2020improved, bechtle2021meta, andrychowicz2016learning}, with a fixed base-learning rate $\alpha$ and meta-learning rate $\eta$ equal to $1\mathrm{e}{-3}$. All base networks are initialized via Xavier Glorot uniform initialization \cite{glorot2010understanding}. 

\begin{itemize}[leftmargin=*, itemsep=3pt, topsep=3pt]

    \item \textbf{Sine}: A tabulated regression problem originally proposed in \cite{finn2018learning}, which involves regressing sine waves, where the amplitude and phase of the sinusoids are varied between tasks. The amplitude varies within $[0.2,5.0]$ and the phase varies within $[-\pi,\pi]$. During meta-training time, five sine waves are generated where points are sampled uniformly from $[-2.0, 2.0]$, and at meta-testing time five sine waves are also generated but are uniformly sampled from $[-5.0, 5.0]$. Identical to \cite{bechtle2021meta}, the base network is a simple feed-forward neural network with 2 dense layers with ReLU activation function, using 40 hidden units each. Training occurs over $S_{meta}=500$ and $S_{base}=100$ gradient steps with a fixed batch size of $100$. 
      
    \item \textbf{MNIST}: An image classification problem originally presented in \cite{lecun1998gradient} that involves classifying images of hand-drawn numeric digits. Identical to \cite{bechtle2021meta}, we partition the original dataset into five separate binary classification tasks (to simulate a prototypical meta-learning setup) where $\Dataset_{Train}$ contains 1 task and $\Dataset_{Test}$ contains 4. The base network is chosen to be the well-known \textit{LeNet-5} convolutional neural network architecture, which is trained over $S_{meta}=1,000$ and $S_{base}=250$ gradient steps respectively with a batch size of $128$. 
      
    \item \textbf{CIFAR-10}: An image classification problem taken from \cite{krizhevsky2009learning}, containing images from 10 different classes. Analogous to the previous dataset, we partition the problem into two separate multi-class classification tasks for meta-training and meta-testing, respectively, where $\Dataset_{Train}$ and $\Dataset_{Test}$ contain 5 distinct classes each. The base network is a convolutional neural network with the following architecture: 5x5 convolution with 32 filters, max pooling, batch normalization, 5x5 convolution with 64 filters, max pooling, batch normalization, dense layer with 256 nodes, dense layer with 128 nodes. Training occurs over $S_{meta}=1,000$ and $S_{base}=2,000$ gradient steps with a batch size of $256$. 
    
    \item \textbf{Surname}: A character-level text recognition problem taken from \cite{rao2019natural}, where the objective is to classify the nationality of surnames from 18 classes. We partition the problem into two separate multi-class classification tasks, where $\Dataset_{Train}$ and $\Dataset_{Test}$ contain 9 distinct classes each. The base network is a recurrent neural network with the following architecture: an embedding layer with an output dimension of 256, an LSTM layer with 64 units, and a dense layer with 256 nodes. Training occurs over $S_{meta}=1,000$ and $S_{base}=2,000$ gradient steps with a batch size of $256$. 
      
\end{itemize}

\section{Results and Analysis}\label{section:c3-results}

\begin{table*}[]
    \centering
    \captionsetup{justification=centering}
    \caption{Results reporting the mean $\pm$ standard deviation \textit{final inference} performance across 10 independent executions of each algorithm, where the MSE is reported for \textit{Sine}, and the ER for \textit{MNIST}, \textit{CIFAR-10} and \textit{Surname}.}
    \subfloat{{%
    \setlength{\tabcolsep}{13pt}
    \begin{tabular*}{0.95\textwidth}{clcccc}
    \cline{2-6}
    \multicolumn{1}{c}{} & &Sine    & MNIST  & CIFAR-10 & Surname  \\ 
    \cline{2-6}
    \noalign{\vskip\doublerulesep\vskip-\arrayrulewidth}
    \cline{2-6}
    \multirow{5}*{\rotatebox{90}{(a) \textit{Training Tasks}}}
       & Baseline           & 3.0280$\pm$1.0911 & 0.0414$\pm$0.0029 & 0.0654$\pm$0.0079 & 0.4224$\pm$0.0930 \\  
       & Random             & 1.4115$\pm$0.7688 & 0.0560$\pm$0.0232 & 0.0642$\pm$0.0301 & 0.3197$\pm$0.0315 \\
       & GP-LFL             & 1.2844$\pm$0.8155 & 0.0387$\pm$0.0278 & 0.0619$\pm$0.1013 & 0.2005$\pm$0.0944 \\ 
       & ML$^3$ Supervised  & 2.1073$\pm$0.7500 & 0.0215$\pm$0.0054 & 0.0323$\pm$0.0099 & 0.2410$\pm$0.0237 \\ \cline{2-6}
       & EvoMAL             & \textbf{1.2670$\pm$0.8052}  & \textbf{0.0056$\pm$0.0009} & \textbf{0.0019$\pm$0.0021} & \textbf{0.1405$\pm$0.0162} \\  
    \cline{2-6}
    \noalign{\vskip\doublerulesep\vskip-\arrayrulewidth}
    \cline{2-6}
    \end{tabular*}%
    }}
    \qquad
    \subfloat{{%
    \setlength{\tabcolsep}{13pt}
    \begin{tabular*}{0.95\textwidth}{clcccc}
    \cline{2-6}
    \multicolumn{1}{c}{} & &Sine    & MNIST  & CIFAR-10 & Surname  \\ 
    \cline{2-6}
    \noalign{\vskip\doublerulesep\vskip-\arrayrulewidth}
    \cline{2-6}
    \multirow{5}*{\rotatebox{90}{(b) \textit{Testing Tasks}}}
       & Baseline           & 4.0735$\pm$1.5581 & 0.0258$\pm$0.0132 & 0.0340$\pm$0.0137 & 0.2025$\pm$0.0231 \\ 
       & Random             & 3.8963$\pm$2.3903 & 0.0592$\pm$0.0516 & 0.0352$\pm$0.0231 & 0.1970$\pm$0.0611 \\ 
       & GP-LFL             & 3.3212$\pm$1.4041 & 0.0265$\pm$0.0286 & 0.0407$\pm$0.0741 & 0.1714$\pm$0.1168 \\ 
       & ML$^3$ Supervised  & 3.5872$\pm$1.8257 & 0.0153$\pm$0.0080 & 0.0149$\pm$0.0041 & 0.1608$\pm$0.0283 \\ \cline{2-6}
       & EvoMAL             & \textbf{3.3099$\pm$1.3685}  & \textbf{0.0053$\pm$0.0028} & \textbf{0.0006$\pm$0.0008} & \textbf{0.0921$\pm$0.0119} \\  
    \cline{2-6}
    \noalign{\vskip\doublerulesep\vskip-\arrayrulewidth}
    \cline{2-6}
    \end{tabular*}
    }}
    \label{table:model-performance}%
\end{table*}

\begin{figure*}[]
    \centering
    \vspace{-6mm}
    \subfloat[\centering Sine]{{\includegraphics[width=0.2\textwidth]{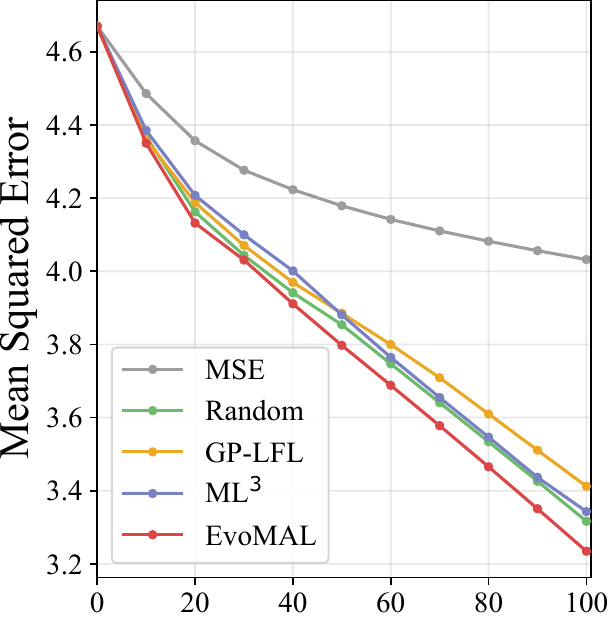}}}%
    \qquad
    \subfloat[\centering MNIST]{{\includegraphics[width=0.2\textwidth]{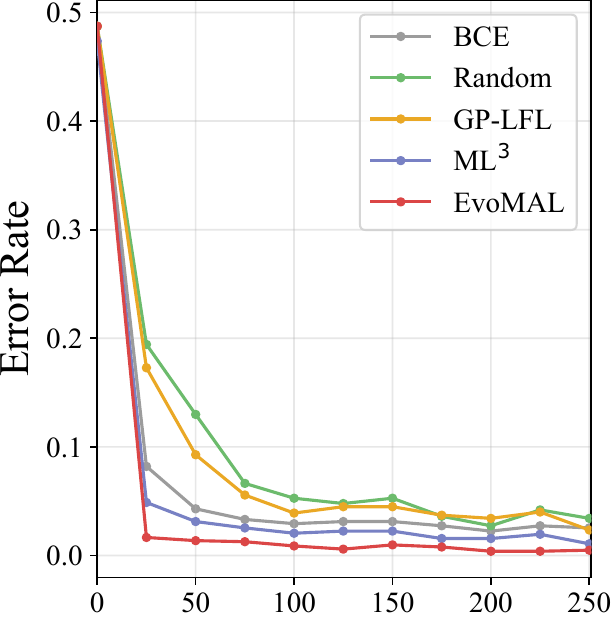}}}%
    \qquad
    \subfloat[\centering CIFAR-10]{{\includegraphics[width=0.2\textwidth]{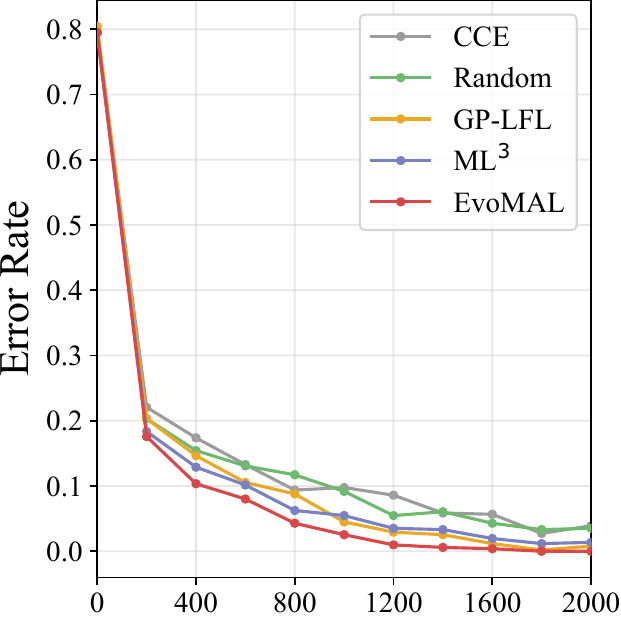}}}%
    \qquad
    \subfloat[\centering Surname]{{\includegraphics[width=0.2\textwidth]{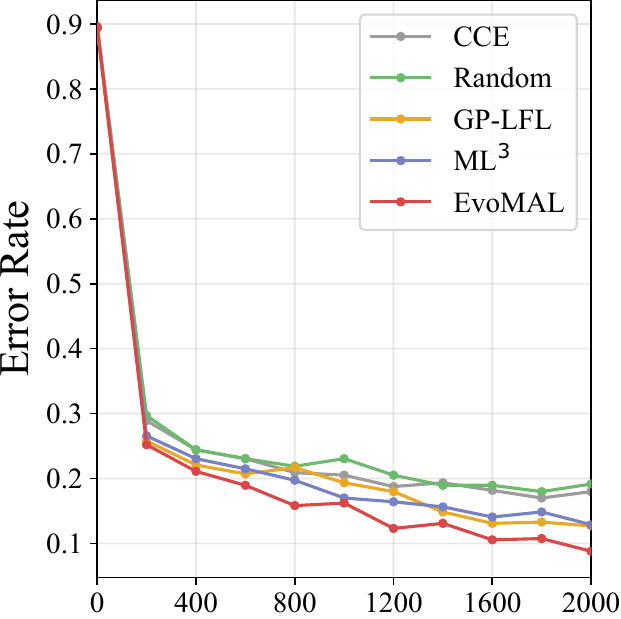}}}%
    \qquad
    \vspace{-2mm}
    \captionsetup{justification=centering}
    \caption{Mean \textit{meta-testing} learning curves on the \textit{out-of-sample} testing tasks across 10 independent executions of each algorithm, showing the performance (y-axis) against gradient steps (x-axis). Best viewed in color.}
    \label{fig:meta-testing-learning-curves}%
\end{figure*}

We present the final inference performance at meta-testing time using the different benchmark methods in Table \ref{table:model-performance}, where (a) presents the in-sample performance on the meta-training tasks and (b) presents the out-of-sample performance on the meta-testing tasks. Analyzing the quantitative results, it is shown that EvoMAL consistently achieves significantly better final inference performance on each of the tested datasets, with much lower $\small MSE \normalsize$s on Sine, and notably lower $\small ER \normalsize$s on MNIST, CIFAR-10, and Surname. The strong in-sample performance demonstrates the effectiveness of EvoMAL in the single-task learning regime, while the out-of-sample performance successfully illustrates the high generalization capabilities when extended to new unseen tasks at meta-testing time. 

\subsection{Comparisons with Benchmark Methods}

In most cases, a clear improvement is observed by using loss function learning techniques, strongly motivating the use of learned loss functions in favor of handcrafted loss functions. Concerning random search, improved performance is achieved on Sine and Surname, similar performance on CIFAR-10 and worse on MNIST compared to the baseline. These results suggest that with the dense symbolic search space containing many promising loss functions, improved search mechanisms are required to achieve better results. The later results by GP-LFL and ML$^3$ empirically confirms this and shows that there is a sufficient exploitable structure in this optimization problem to motivate the design of more sophisticated loss function learning techniques.

Contrasting the performance of EvoMAL to its derivative methods GP-LFL and ML$^3$, notable gains in inference performance are shown on all the tested datasets. These results reveal two key findings: first, further evidence that GP-based methods benefit significantly from introducing local search mechanisms, and second that gradient-based methods can successfully be utilized to achieve this task in a computationally tractable way. Our experiments were all conducted on a single consumer-level GPU, in contrast to \cite{gonzalez2020improved}, which required a supercomputer and significantly reduced values of $S_{base}$ at meta-training time.

\subsection{Convergence and Sample-Efficiency}

\begin{figure*}[htb]
    \centering
    \subfloat[\centering Sine]{{\includegraphics[width=0.2\textwidth]{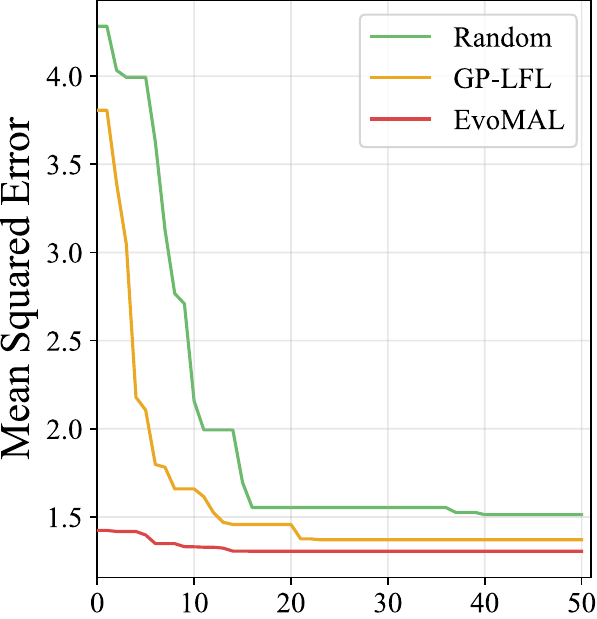}}}%
    \qquad
    \subfloat[\centering MNIST]{{\includegraphics[width=0.2\textwidth]{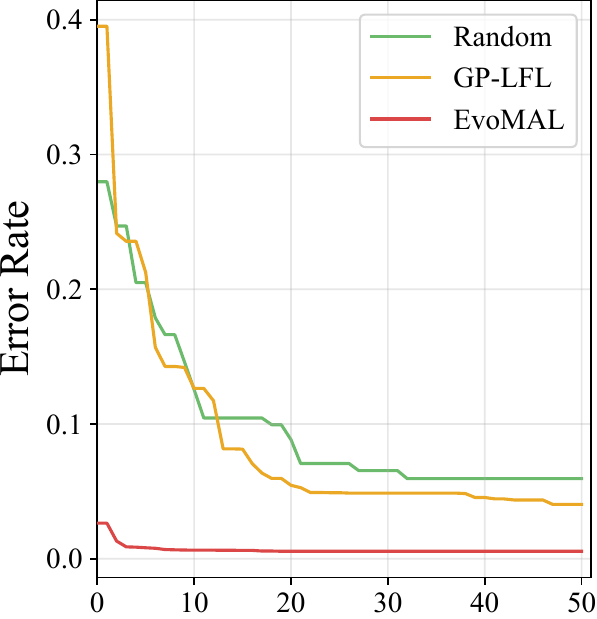}}}%
    \qquad
    \subfloat[\centering CIFAR-10]{{\includegraphics[width=0.2\textwidth]{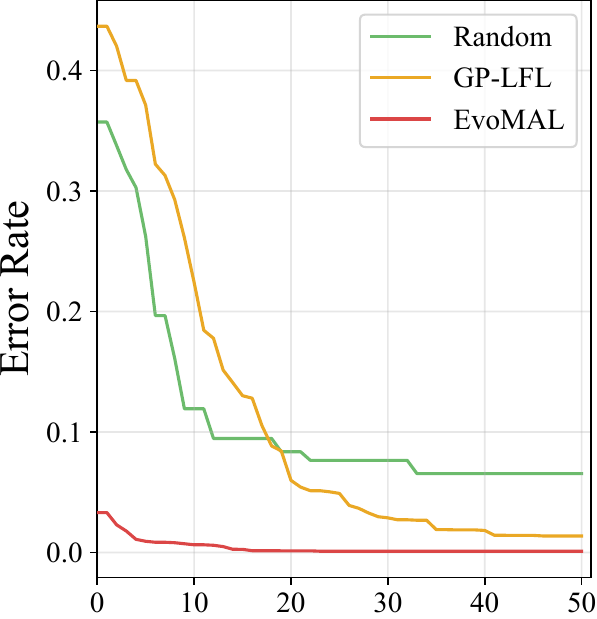}}}%
    \qquad
    \subfloat[\centering Surname]{{\includegraphics[width=0.2\textwidth]{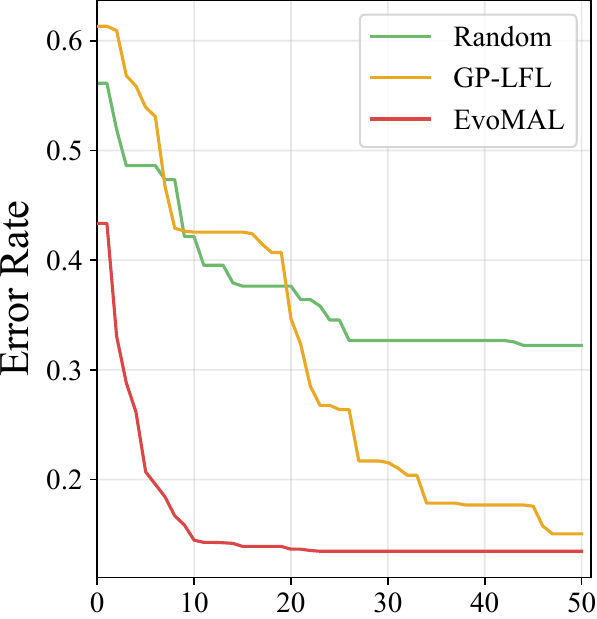}}}%
    \qquad
    \captionsetup{justification=centering}
    \caption{Mean \textit{meta-training} learning curves across 10 independent executions of each algorithm, showing the fitness score (y-axis) against outer iterations (x-axis), where an iteration is equivalent to 25 evaluations. Best viewed in color.}
    \label{fig:meta-training-learning-curves}%
\end{figure*}

To further discern the benefits of using the learned loss functions produced by EvoMAL, the meta-testing learning curves on the in-sample and out-of-sample tasks are presented in Figure \ref{fig:meta-testing-learning-curves}. Examining the qualitative results, it is observed that EvoMAL consistently produces improved performance compared to the benchmark methods, and it does so in far fewer gradient steps at all stages in the learning process, demonstrating improved sample efficiency and convergence capabilities, which is a desirable characteristic in regimes where either data or computational resources are low. Furthermore, the performance generally shows that EvoMAL only requires a small subset of the total gradient steps to surpass the final performance of the baseline.

\subsection{Meta-Training Dynamics}

To analyze the search effectiveness of EvoMAL the meta-training learning curves comparing the search performance (as quantified by the fitness function) at each iteration are given in Figure \ref{fig:meta-training-learning-curves}. Based on the results, it is evident that adding local search mechanisms into the EvoMAL framework dramatically reduces the total number of iterations needed to find performant loss functions compared to random search and GP-LFL. Furthermore, while EvoMAL often finds comparable fitness loss functions to GP-LFL after 50 iterations, the performance when evaluated, \textit{i.e.} at meta-testing time, corresponds to a better generalizing loss function compared to GP-LFL as shown in Table \ref{table:model-performance}.

\subsection{Loss Function Paramaterization}

\begin{table}[]
\renewcommand{\arraystretch}{1.3}
\centering
\captionsetup{justification=centering}
\caption{Contrasting the mean $\pm$ standard deviation number of trainable loss parameters $\phi$ between ML$^3$ and EvoMAL.}
\begin{tabular}{lcccc}
\hline
            & Sine               & MNIST             & CIFAR-10          & Surname           \\ \hline \hline
ML$^3$      & 2,650    & 2,650   & 2,650  & 2,650  \\ 
EvoMAL      & 30.9 $\pm$ 10.3        & 30.8 $\pm$ 6.8   & 26.8 $\pm$ 9.5   & 28.3 $\pm$ 13  \\ \hline \hline
\end{tabular}
\label{table:parameter-comparison}%
\vspace{-5pt}
\end{table}

In addition to the performance benefits of EvoMAL, a compelling finding is that compared to its gradient-based derivative method, ML$^3$, only a small fraction of the number of trainable loss parameters $\phi$ is required, as shown in Table \ref{table:parameter-comparison}, where the meta-loss networks in ML$^3$ use a feed-forward neural network with two input nodes followed by two dense layers of 50 nodes each and one output node \cite{bechtle2021meta}. These results show that in ML$^3$, the meta-loss networks are significantly over-parameterized, as less than $\sim$1-2\% of the total number of trainable loss parameters $\phi$ are needed in the loss functions learned by EvoMAL compared to that of ML$^3$ across all the tested datasets. Consequently, the loss functions learned by EvoMAL have improved inference speed, \textit{i.e.} reduced cost to compute loss values at meta-testing time, due to not having to propagate through so many parameters.

\section{Conclusions and Future Work}\label{section:c3-summary}

This work presents a new framework for meta-learning symbolic loss function via a hybrid neuro-symbolic search approach called Evolved Model-Agnostic Loss (EvoMAL). The proposed method poses the problem of loss function learning in terms of a bilevel optimization problem, where the outer optimization problem involved learning a set of symbolic loss functions via GP, and the inner optimization problem consisted of learning the weights of the parameterized loss functions found in the outer optimization process via unrolled differentiation. Integration of the outer and inner optimization problems was performed seamlessly by introducing a linear time transition procedure converting the GP expression tree-based loss functions into trainable meta-loss networks. 

Our analysis of the learned loss functions produced by the newly proposed framework shows several benefits compared to handcrafted loss functions, and state-of-the-art loss function learning techniques. Empirical results show improved inference performance, convergence, and sample efficiency. Furthermore, this performance can successfully generalize to new unseen tasks not seen at meta-training time. Unlike existing methods for loss function learning, the proposed framework can be combined with any model representation amenable to a gradient-descent style training procedure for any supervised learning task, due to the generality in the search space design. Additionally, EvoMAL is the first GP-based loss function learning approach to integrate local search mechanisms into the learning process in a computationally feasible manner.

Despite the effectiveness of EvoMAL, there are still aspects of the framework that can be further improved upon and developed in future work. Firstly, we posit that introducing rejection protocols that filter out non-promising or gradient-equivalent loss functions similar to what was proposed in \cite{liu2021loss} and \cite{li2021autoloss}, can reduce the number of evaluations required, thus reducing the runtime further. In addition, investigating alternative meta-optimization strategies such as implicit differentiation \cite{lorraine2020optimizing} or first-order gradient-based alternatives \cite{nichol2018first} to unrolled differentiation is a promising area to explore, since a computational bottleneck in EvoMAL, and its derivative method ML$^3$ is 
having to differentiate through the optimization path.

\bibliographystyle{ACM-Reference-Format}
\bibliography{references}

\end{document}